\documentclass{article}

\PassOptionsToPackage{numbers, compress}{natbib}
\usepackage[preprint]{neurips_2025}

\usepackage[utf8]{inputenc} 
\usepackage[T1]{fontenc}    
\usepackage{hyperref}       
\usepackage{url}            
\usepackage{booktabs}       
\usepackage{amsfonts}       
\usepackage{nicefrac}       
\usepackage{microtype}      

\usepackage{graphicx}
\usepackage{subcaption}
\usepackage{amsmath}
\usepackage{adjustbox}
\usepackage{multirow}
\usepackage{pifont}

\usepackage[capitalize]{cleveref}
\crefname{section}{Sec.}{Secs.}
\Crefname{section}{Section}{Sections}
\Crefname{table}{Table}{Tables}
\crefname{table}{Tab.}{Tabs.}

\usepackage[dvipsnames]{xcolor}

\definecolor{c1}{HTML}{765D97} 
\definecolor{c2}{HTML}{006400}
\definecolor{c3}{HTML}{fc6160}
\definecolor{myblue}{HTML}{E6F3FC} 
\definecolor{mygray}{HTML}{DBE2E9} 
\definecolor{mygreen}{HTML}{006400} 

\hypersetup{
    colorlinks=true,
    linkcolor=black,
    urlcolor=c1,
    citecolor=c2,
}

\title{SORCE: Small Object Retrieval in Complex Environments}

\author{%
  \hspace{-8mm} Chunxu Liu$^{1,2}$\thanks{Equal Contribution. Work is done during internship at Sensetime.} \quad Chi Xie$^{2, 3*}$ \quad Xiaxu Chen$^{2, 4}$ \quad Wei Li$^2$ \quad Feng Zhu$^2$ \quad Rui Zhao$^2$\quad Limin Wang$^{1,5}\thanks{Corresponding author  (lmwang@nju.edu.cn).}$ \\
  $^1$State Key Laboratory for Novel Software Technology, Nanjing University\\
  $^2$Sensetime Research \quad $^3$ Tongji University \quad $^4$ Beijing Institute of Technology \quad $^5$Shanghai AI Lab\\
  \newline
  \\
  \hspace{-10mm} \textbf{\url{https://github.com/MCG-NJU/SORCE}}
}

\begin{document}

\maketitle
\begin{abstract}
Text-to-Image Retrieval (T2IR) is a highly valuable task that aims to match a given textual query to images in a gallery. Existing benchmarks primarily focus on textual queries describing overall image semantics or foreground salient objects, possibly overlooking inconspicuous small objects, especially in complex environments. Such small object retrieval is crucial, as in real-world applications, the targets of interest are not always prominent in the image. Thus, we introduce \textbf{SORCE} (\textbf{S}mall \textbf{O}bject \textbf{R}etrieval in \textbf{C}omplex \textbf{E}nvironments), a new subfield of T2IR, focusing on retrieving small objects in complex images with textual queries. We propose a new benchmark, SORCE-1K, consisting of images with complex environments and textual queries describing less conspicuous small objects with minimal contextual cues from other salient objects. Preliminary analysis on SORCE-1K finds that existing T2IR methods struggle to capture small objects and encode all the semantics into a single embedding, leading to poor retrieval performance on SORCE-1K. 

Therefore, we propose to represent each image with multiple distinctive embeddings. We leverage Multimodal Large Language Models (MLLMs) to extract multiple embeddings for each image instructed by a set of Regional Prompts (ReP). Experimental results show that our multi-embedding approach through MLLM and ReP significantly outperforms existing T2IR methods on SORCE-1K. Our experiments validate the effectiveness of SORCE-1K for benchmarking SORCE performances, highlighting the potential of multi-embedding representation and text-customized MLLM features for addressing this task.

\end{abstract}

\section{Introduction}
\label{sec:intro}

\begin{figure}[t]
    \centering
    \includegraphics[width=0.85\linewidth]{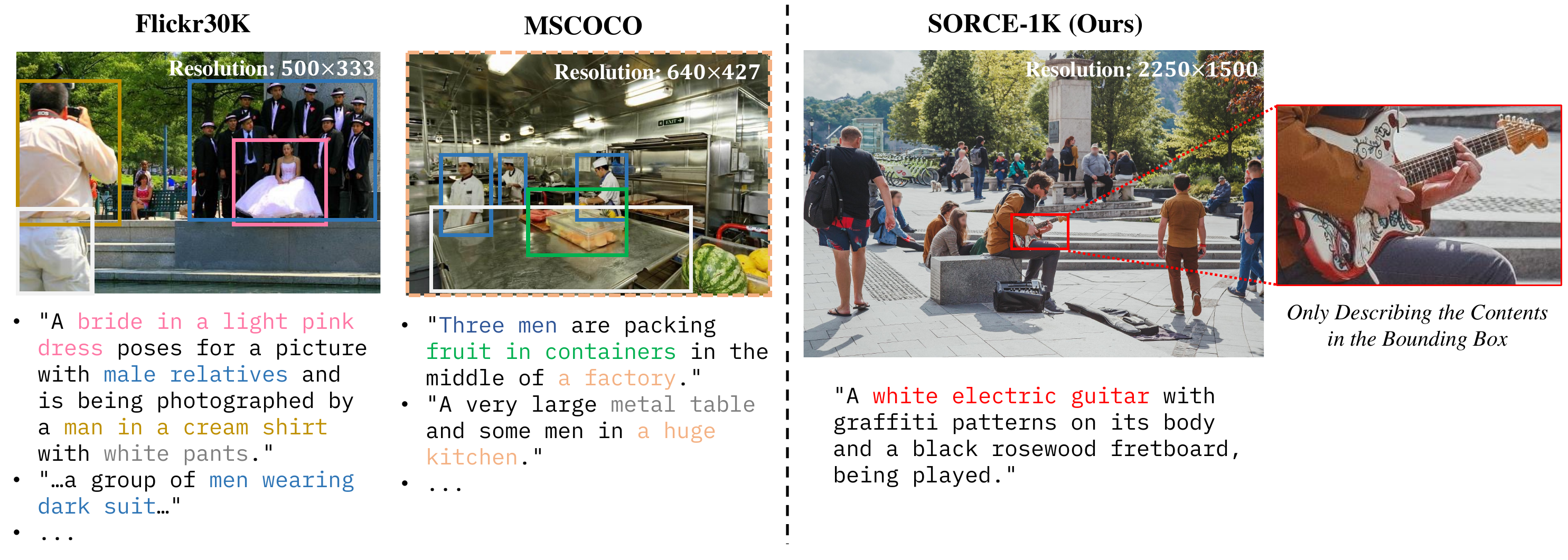}
    \caption{
    \textbf{Instances from different retrieval benchmarks.}
    We draw the bounding boxes of the referred objects in the image.}
    \vspace{-0.5cm}
    \label{fig:introfig}
\end{figure}

While traditional Text-to-Image Retrieval (T2IR) focuses on retrieving images that match a given textual query, an important yet often overlooked challenge is retrieving small objects within complex scenes. In many real-world applications, the target of interest is not the dominant subject, but a small, inconspicuous object in a cluttered background. Despite significant advancements in T2IR methods~\cite{clip, siglip, vista, longclip}, existing frameworks struggle with such fine-grained retrieval, highlighting the need for dedicated research.

The challenge of retrieving small objects in complex environments with textual queries is highly relevant to various real-world applications. In video surveillance, it is essential for discovering inconspicuous yet critical items, such as concealed weapons in crowded areas or abandoned luggage in transportation hubs; in smart city systems, it helps detect traffic violations (e.g., unlicensed vehicles) and monitor infrastructure integrity (e.g., cracks in bridges). In cybersecurity, it can help discover tampered watermarks and manipulated content; in consumer applications, it enables users to locate specific small objects within personal photo collections efficiently.

Despite the importance of small object retrieval, existing T2IR benchmarks primarily focus on retrieving images based on holistic descriptions rather than identifying fine-grained, non-salient objects. Popular T2IR benchmarks like Flickr30K~\cite{flickr} and COCO~\cite{coco} highlights the main object or the whole picture in the image. Other datasets with image-text pairs~\cite{docci,sharegpt4v} provide more detailed descriptions, but still focus on describing the full image rather than specific small objects.

To address this gap, we introduce \textbf{S}mall \textbf{O}bject \textbf{R}etrieval in \textbf{C}omplex \textbf{E}nvironments (\textbf{SORCE}), a task focusing on retrieving the target image containing a specific small object.
Here, \textit{small} objects refer to the objects that are \textit{visually} and \textit{resolution-wise} small in the image, not semantically small.

For this task, we construct the SORCE-1K benchmark by collecting images from the SA-1B dataset~\cite{sam}. As illustrated in \cref{fig:introfig}, it is a high-resolution segmentation dataset with 1B mask annotations across 11M images, guaranteeing a high level of environment complexity. For each image, we identify a small target object and carefully craft a descriptive caption that uniquely specifies it, minimizing ambiguity within the benchmark. The resulting dataset, \textbf{SORCE-1K}, is a comprehensive benchmark for evaluating small object retrieval in complex environments.

In image retrieval, candidate images are typically pre-encoded into features, and retrieval is performed based on feature similarity with the query. Consequently, the image feature extraction is independent of the query content. This presents a significant challenge for feature extraction to capture every fine-grained detail within a single feature representation, making small object retrieval particularly difficult. Especially, the objects of interest in our SORCE-1K are often non-salient and embedded in complex scenes. Thus, relying on a single image feature for SORCE is likely unsatisfying. 

To tackle the challenges of the SORCE task, we propose a simple yet effective starting point: representing an image with multiple features instead of a single embedding. Conventional feature extractors tend to prioritize the most salient objects, often overlooking fine-grained details. Extracting multiple feature representations can help mitigate this issue by capturing different aspects of an image. This can be naturally achieved using feature extractors based on Multimodal Large Language Models (MLLMs). Unlike traditional encoders, MLLMs can generate different feature embeddings based on different prompts~\cite{prior1,prior2,gme}, allowing them to focus on different aspects of the image while preserving global context. By leveraging this property, we can obtain a more comprehensive representation that improves retrieval performance for small, non-salient objects in complex environments.


Our main contributions are summarized as follows:
\begin{enumerate}
    \item We introduce SORCE, a new subfield of Text-to-Image Retrieval (T2IR) that addresses the challenge of retrieving small, less conspicuous objects within images with complex environments based on textual queries.
    \item We develop a new benchmark, SORCE-1K, for facilitating research on the SORCE task. SORCE-1K comprises 1,023 carefully curated images with complex backgrounds and textual queries that describe less prominent small objects with minimum surrounding context. 
    \item We propose an MLLM-based multi-embeddings approach, with Regional Prompts (ReP) to extract multiple image features focusing on different aspects. Superior performances over existing T2IR methods on SORCE-1K highlight the potential of multi-embedding representation and text-guided MLLM features for addressing the SORCE task.
    
    
\end{enumerate}

\section{Related Work}
\label{sec:related_work}

\subsection{Text-to-Image Retrieval Benchmarks}

Text-to-Image Retrieval (T2IR) aims to retrieve relevant images from a candidate pool based on a given textual query. In practice, the candidate pool consists of pre-extracted image features, agnostic of the query content. The common approach is to first extract the text query embedding, then compute the cosine similarity between this text embedding and the embeddings of the candidate images. The top-ranked images with the highest similarity scores are retrieved as the final results.

Common benchmarks such as Flickr30K~\cite{flickr} and COCO~\cite{coco} provide five brief captions for each image, describing the main object or event. Other datasets such as ShareGPT4V~\cite{sharegpt4v} and DOCCI~\cite{docci} further extend the captions of images to be more detailed descriptions. Urban-1K~\cite{urban} is another retrieval dataset for urban images with detailed descriptions. However, all of the above retrieval benchmarks are dedicated to enriching the query to be a detailed description of a whole image or the obvious objects in the image. In contrast, our benchmark SORCE-1K introduces a novel setting, which only contains the minimum required context in the query description. 
This approach aims to prevent retrieval models from taking shortcuts by recognizing and relying on non-target contextual descriptions. Therefore, our benchmark supports the robust retrieval evaluation of the specified target.

\subsection{Visual Discovery of Small Objects}
The visual discovery of small objects remains a longstanding challenge in computer vision, intersecting with several related domains , including the following. \textbf{I2I retrieval.} Handling small objects in I2I retrieval presents significant challenges, as highlighted in~\cite{sattler2012image}. Prior works have addressed this by improving traditional or CNN-based descriptors~\cite{iscen2017efficient, chen2019learning}. However, these methods are limited to single-modality retrieval and are not aligned with textual semantics. \textbf{Small object detection/segmentation.}
Most small object detection or segmentation methods focus on a limited set of classes~\cite{li2017perceptual, noh2019better, yang2022querydet, wang2019miss, sang2022small}. Open-vocabulary approaches~\cite{minderer2023scaling, zhang2023simple} still rely on a predefined set of target objects. Even advanced models like SAM~\cite{sam} and DINO-X~\cite{ren2024dinoxunifiedvisionmodel}, which can detect objects without text prompts, often over-segment the image and are not suitable for our SORCE task.
\textbf{Fine-grained VQA.}
MLLMs\cite{llava,llava1.5,llava-next} have been increasingly used to tackle fine-grained visual understanding. Benchmarks like SEED\cite{SEED}, MME~\cite{MME}, and MMBench~\cite{MM-Bench} test models on tasks involving spatial reasoning, attribute comparison, object recognition, and spotting subtle differences. V*~\cite{vstar} introduces a visual search task that requires grounding small objects in high-resolution images based on questions, pushing MLLMs to identify fine-grained details progressively. Unlike these VQA-based approaches, SORCE evaluates fine-grained perception through retrieval, where the model is not given any prior knowledge of the target object. This provides a complementary view of the model’s ability to localize and recognize small details without explicit textual input.

\subsection{MLLM Embeddings}


Recently, a new line of research has explored the use of MLLMs as feature extractors~\cite{E5-V}. Trained with next-word prediction loss, MLLMs can effectively process and integrate visual and textual information. E5-V~\cite{E5-V} first discovered that, when provided with carefully crafted prompts under the ``one-word limitation" (e.g., ``\texttt{<image> Summarize the above image in one word:}"), MLLMs leverage their multimodal comprehension to predict a one-word summary in the final token, effectively condensing input tokens into a compact embedding, and eliminating the feature gap between different modalities. Building on this insight, subsequent studies have shown that MLLMs have significant potential to overcome the limitations of conventional vision-language models (VLMs) and achieve outstanding performance in multimodal retrieval tasks~\cite{lamra, gme, discrim, cirr, uniir}.

In this work, we extend these findings by demonstrating that MLLMs can extract text-guided features through simple prompting. We highlight their potential for small object retrieval in complex scenes, leveraging their instruction-following ability.

\section{SORCE-1K Benchmark}
\label{sec:benchmark}

\subsection{Dataset Highlights and Statistics}
The proposed benchmark follows the same task formulation and evaluation protocol as existing text-to-image retrieval benchmarks. However, its key challenges differ from traditional retrieval benchmarks, as it emphasizes retrieving the image that contains a specific small target in potentially complex environments. Below, we highlight the key characteristics of this benchmark.

\begin{figure}
    \centering
    \includegraphics[width=0.8\linewidth]{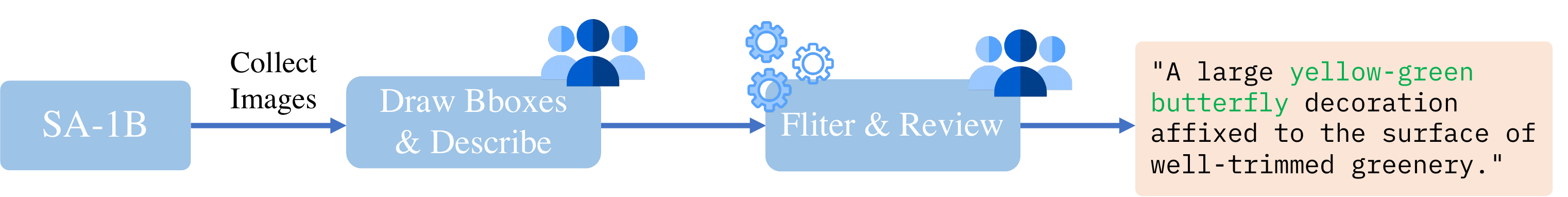}
    \caption{\textbf{The construction process for SORCE-1K.}}
    \label{fig:bench-process}
\end{figure}

\begin{figure}
    \centering
    \includegraphics[width=0.7\linewidth]{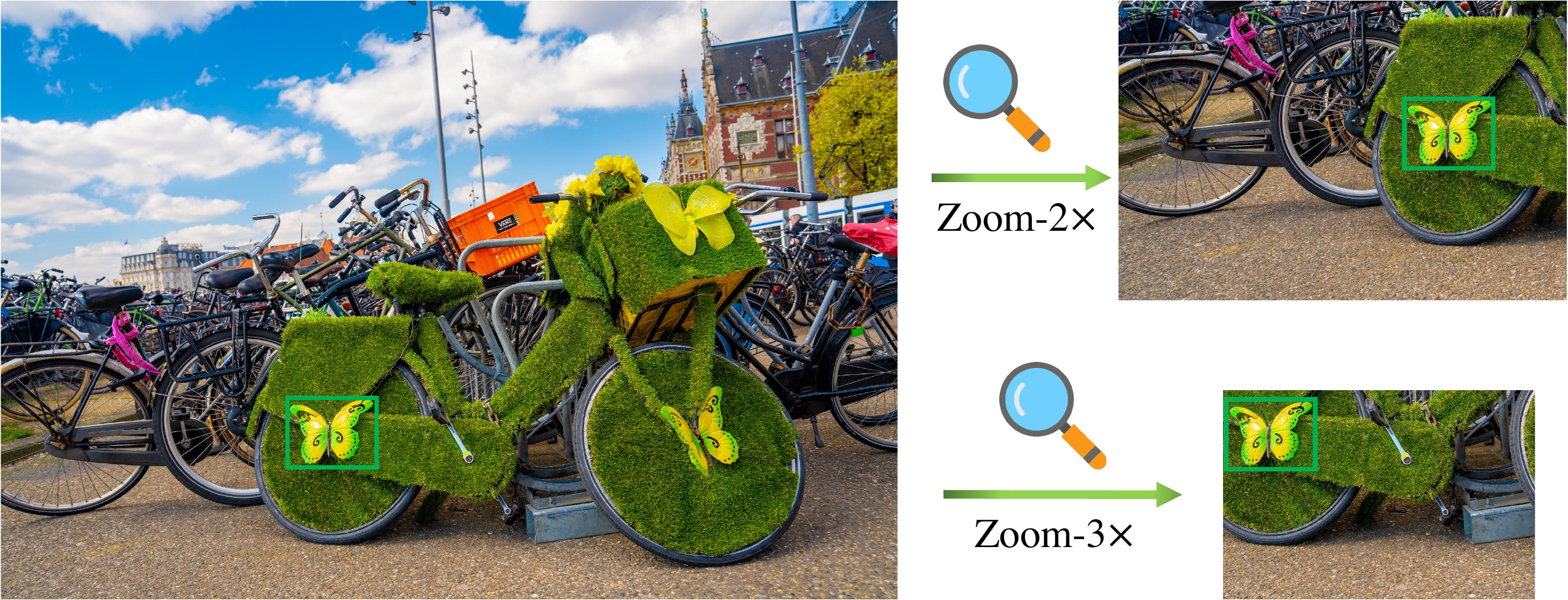} 
    \caption{\textbf{Example of the difficulty levels.} We perform zooming $2\times$ and $3\times$ and crop the image while ensuring the object is in the resulting frame. More qualitative examples in~\cref{supp:more_qualitative}.}
    \vspace{-0.5cm}
    \label{fig:bench-vis}
\end{figure}

\noindent \textbf{Small targets.}
SORCE-1K requires a model to retrieve an image containing a specific small target object. The target is typically a less salient element within the image, often blending into the background or hidden among details.
This presents a greater challenge compared to traditional retrieval settings, where the target is usually prominent foreground objects or the whole scene.

\noindent \textbf{Complex environments.}
The targets are usually embedded in complex scenes, often appearing as background components or fine details.  
The scene complexity is ensured by using SA-1B~\cite{sam} as the image source and selecting only images with a large number of masks.
This requires the model to recognize and differentiate between entire scenes and the objects within them.  
Note that the numbers of masks in images in our benchmark are presented in~\cref{supp:environment_complexity}.

\noindent \textbf{Local and detailed descriptions.}
Each target is referenced by a detailed description that uniquely matches it without overlapping with objects in other images from the dataset.  
The corresponding description primarily focuses on local features of the target object, rather than its surrounding context or the entire scene.
This ensures that the retrieval model need to identify the target itself rather than relying on contextual cues mentioned in the descriptions as shortcuts.

\noindent \textbf{Multiple difficulty levels.}
Retrieving full-size images can be too challenging.
To facilitate comparison and analysis across different scales, SORCE-1K is designed with two additional ``zoom-in'' settings where the target object is kept while the image context is cropped. The 3 settings (Full Res., Zoom-2$\times$, Zoom-3$\times$) are depicted in~\cref{fig:bench-vis}. This helps understand gaps between traditional retrieval tasks and the proposed task setting.

\noindent \textbf{High quality.}
The entire benchmark is annotated by researchers themselves rather than crowd-sourced annotators. Additionally, measures such as manual checking are implemented during the annotation process to ensure uncompromised quality.

\begin{figure}
    \centering
    \begin{subfigure}{0.4\linewidth}
        \centering
        \includegraphics[width=1.0\linewidth, trim=0cm 1cm 0cm 0.89cm, clip=true]{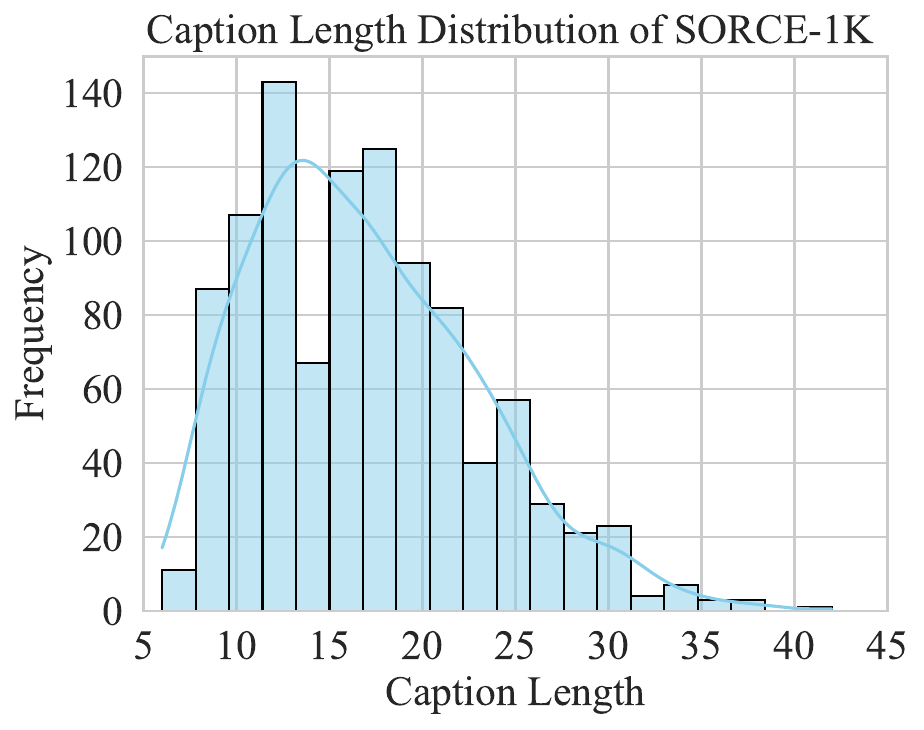}
        \caption{Caption length distribution}
    \end{subfigure}
    \hspace{0.05\linewidth} 
    \begin{subfigure}{0.4\linewidth}
        \centering
        \includegraphics[width=1.0\linewidth, trim=0cm 1cm 0cm 0.89cm, clip=true]{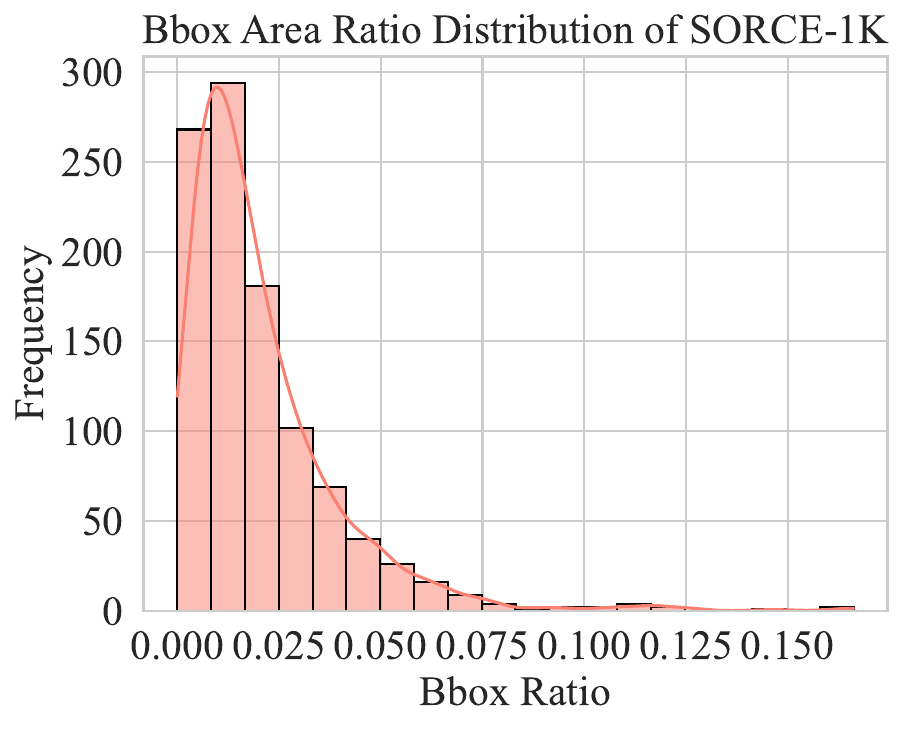}
        \caption{Bbox area ratio distribution}
    \end{subfigure}
    \caption{\textbf{Statistics for the proposed SORCE-1K benchmark.}}
    \label{fig:sorce-stats}
    \vspace{-0.6cm}
\end{figure}

\noindent \textbf{Dataset statistics.}
To construct the dataset, a total of 2.1K images from SA-1B are labeled and examined, and about half are discarded for not meeting these criteria.
The result dataset is an evaluation-only benchmark with 1,023 images. Each image is annotated with one description and one box, though the box is not used for the retrieval task.
The description refers to the object in the box on the image.
As is shown in~\cref{fig:sorce-stats}(a), the descriptions in the dataset are at least 6 words long, at most 42 words, with an average of 16.9 words.
The bounding boxes are very small compared with the high-resolution image. As is shown in~\cref{fig:sorce-stats} (b), the small objects to retrieve usually take up less than 10\% of the images.
The evaluation metric is the same as standard retrieval benchmarks.

\subsection{Dataset Construction}
This benchmark is annotated manually by researchers, followed by automatic and manual verification.

\noindent \textbf{Data source.}
To construct a benchmark featuring small objects in complex environments, we use images from SA-1B~\cite{sam}.
It has (1) high-resolution images, which are more likely to contain small objects, and (2) complex scenes, quantifiable by the number of masks in SA-1B annotations.

\noindent \textbf{Region-description pair annotation.}
For each image, we manually check whether it contains a target that (1) is relatively small compared to the entire image, (2) can be described using natural language, and (3) has a unique combination of local features that distinguishes it from all other instances in the dataset.
If such a target exists, we annotate it with a bounding box and a corresponding unique description; otherwise, we exclude the image from the dataset.

\noindent \textbf{Dataset tiering with difficulty.}
For each Full Res. sample, we divide it by $2\times2$ and $3\times3$ to obtain a Zoom-2$\times$ sample and a Zoom-3$\times$ sample. To ensure the target object remains intact, we select the sub-image that covers the bounding box the most, and extend the region when necessary to fully include the bounding box. The description annotation is shared across all three levels.
Compared to the Full Res. image, the Zoom-2$\times$ setting is easier, as the target is more prominent, while the Zoom-3$\times$ setting is the easiest.

\noindent \textbf{Automatic check.}
We apply a few simple automatic filtering steps to the candidate samples from the previous step. These steps include (1) removing samples where the bounding box is too large and dominant in the image and (2) removing or correcting samples with overly short descriptions, as they are likely too simple or non-unique. These steps help ensure that annotation errors are minimized.

\noindent \textbf{Manual review.}
Finally, we perform a manual review of the annotated dataset. Each image is visualized along with its annotated region-description pair. We examine the visualizations and remove samples where the descriptions are inaccurate or ambiguous, particularly those that may refer to multiple instances in the dataset.
Please see~\cref{supp:construction_details} for more details regarding the steps above.

\subsection{Other Application Scenarios}



\noindent \textbf{Visual grounding and described object detection.}
Since each image is annotated with a single bounding box associated with a language description, the proposed benchmark can serve directly as a visual grounding~\cite{rec} benchmark. It is also applicable to described object detection~\cite{dod}, a related task that involves detecting all possible objects in an image set based on a given language description.

\noindent \textbf{Visual search.}
Additionally, because the target object in each image is relatively small, the benchmark can be adapted for the visual search~\cite{vstar} task, where a model must answer a question about a small target within an image. This can be achieved by prompting an LLM to convert the target object's description into a question-answer pair focused on a specific attribute of the object.

\noindent \textbf{Instance-level retrieval.}
While the proposed benchmark is primarily for image-level retrieval, it is also applicable to instance-level retrieval tasks~\cite{xdetr}. These tasks typically require explicit bounding box predictions, which our benchmark does not inherently demand but provides in the annotations.
\section{Method}
\label{sec:method}

\subsection{Multiple Feature Representation}
As shown in~\cref{tab:sorce}, existing approaches obtain poor results on the Full Res. setting of SCORCE-1K. This shows that compressing all visual detail into a single image feature remains highly challenging for existing approaches either based on VLMs~\cite{clip,EVA} or MLLMs~\cite{llava-next,E5-V}.

\begin{figure}[t]
    \centering
    \vspace{-2cm}
    \begin{subfigure}{1.0\linewidth}
        \centering
        \includegraphics[width=0.7\linewidth, trim=6.5cm 10cm 15cm 9cm, clip=true]{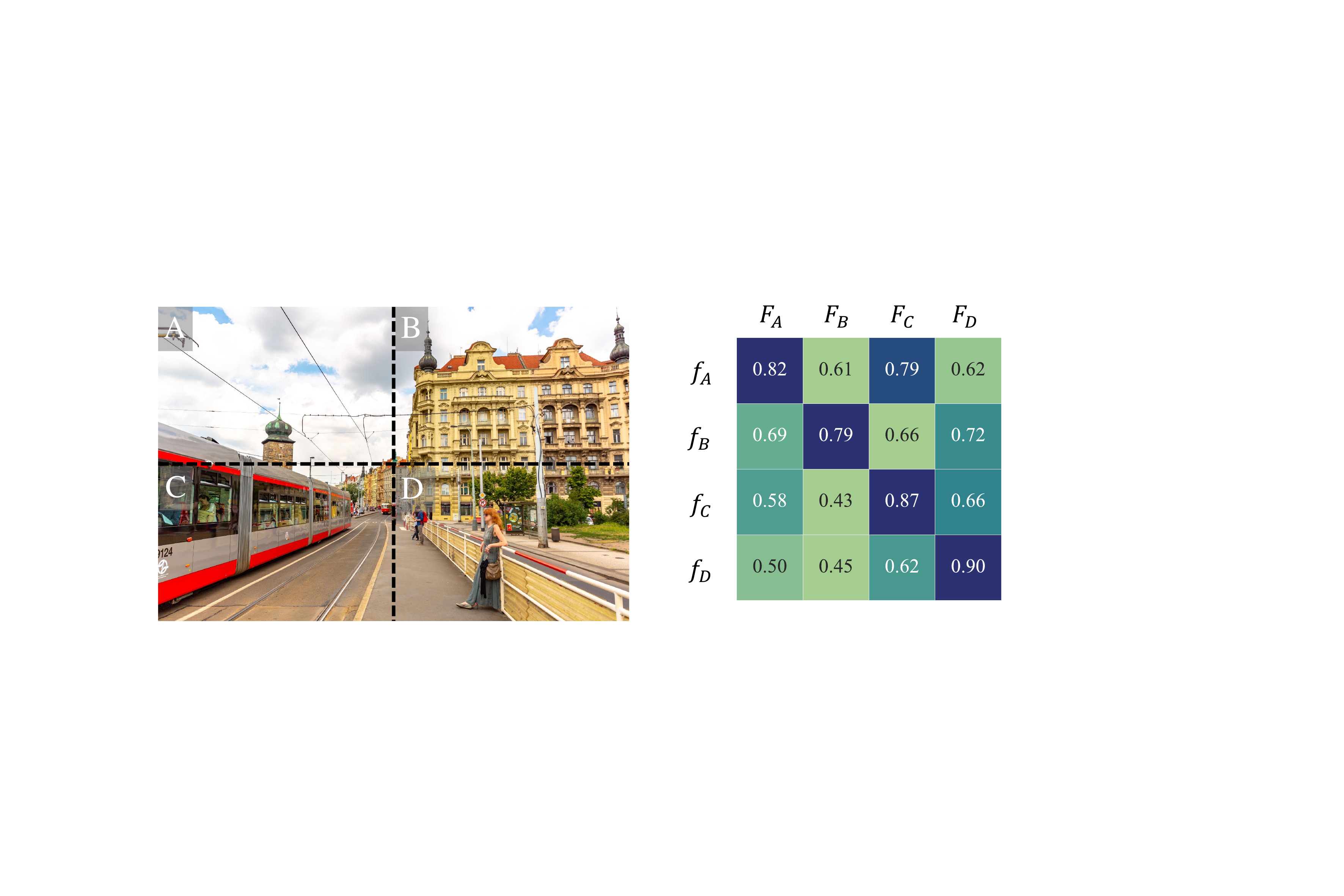}
        \caption{\textbf{Similarity matrix of regional prompts generated features ($f_{A, B, C, D}$) and independently generated features from $2\times2$ Split ($F_{A, B, C, D}$).} Both sets of features are extracted using E5-V. $F_{A, B, C, D}$ are extracted by prompt: ``\texttt{Summarize the above image in one word.}''}
        \vspace{-0.1cm}
        \label{fig:directional-a}
    \end{subfigure}
    \begin{subfigure}{0.4\linewidth}
        \centering
        \includegraphics[width=\linewidth, trim=20cm 15cm 22cm 15cm, clip=true]{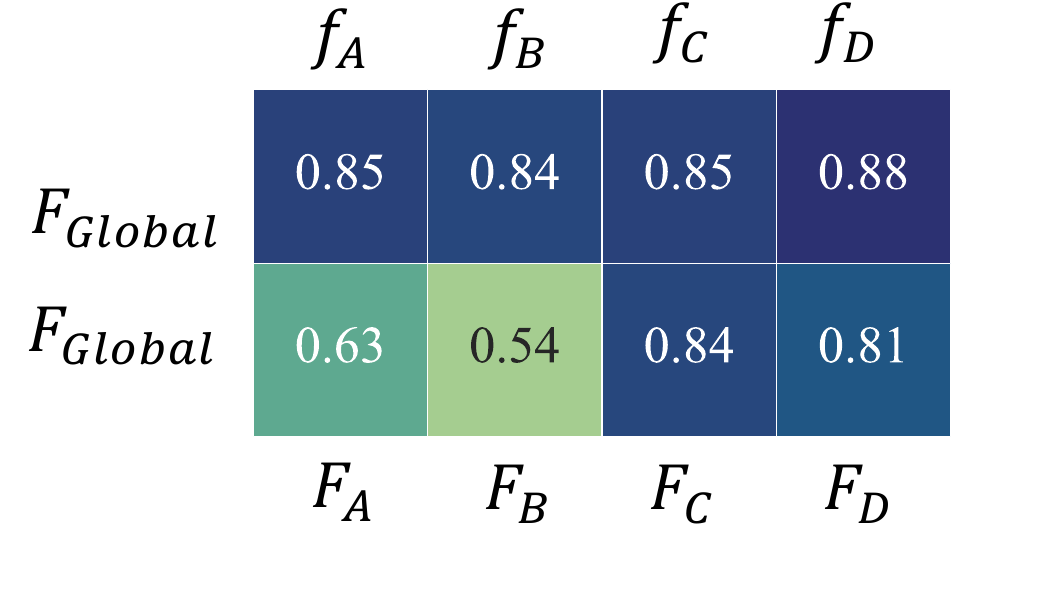}
        \vspace{0.2cm}
        \caption{\textbf{Similarity matrix of global image feature ($F_{global}$) and $f_{A, B, C, D}$, $F_{A, B, C, D}$.} }
        \label{fig:directional-b}
    \end{subfigure}
    \hspace{0.05\linewidth} 
    \begin{subfigure}{0.5\linewidth}
        \centering
        \includegraphics[width=0.7\linewidth, trim=2cm 1.5cm 2cm 2cm, clip=true]{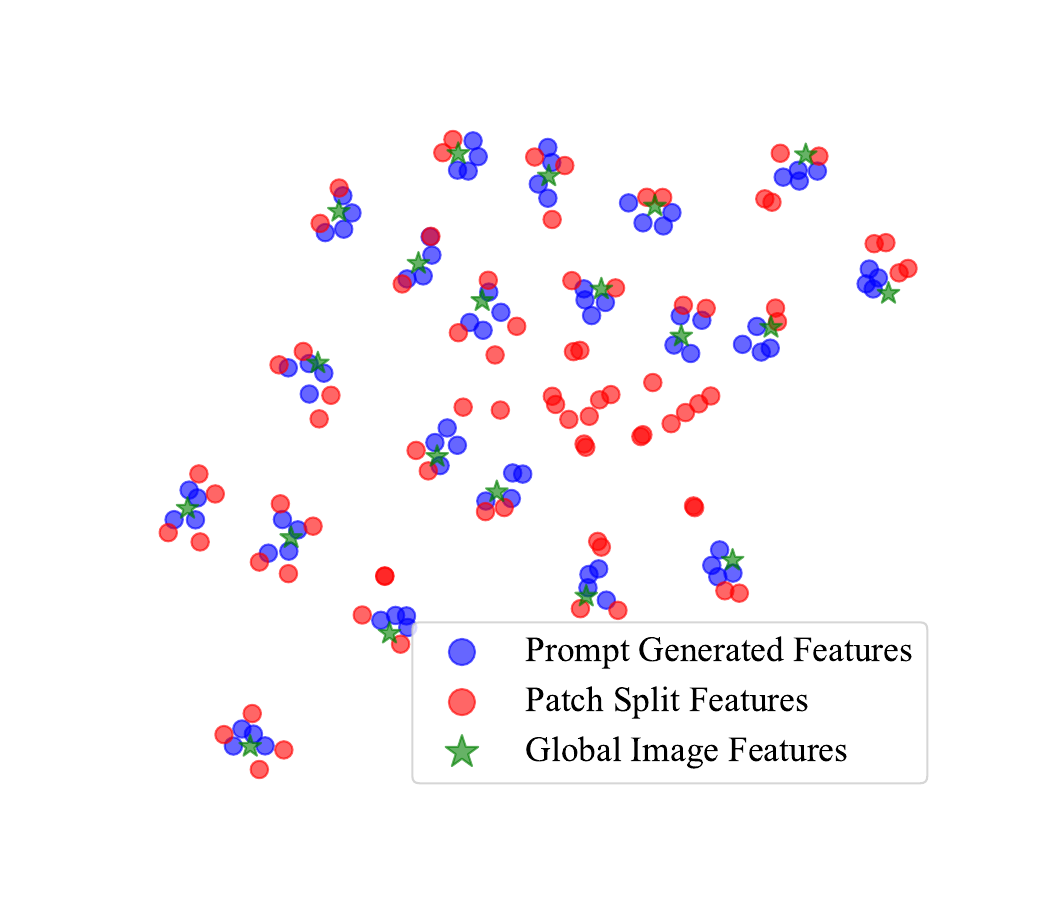}
        \caption{\textbf{Independently generated features from $2\times2$ splits.}
        We randomly select 20 image samples from SORCE-1K.}
        \label{fig:directional-c}
    \end{subfigure}
    \caption{\textbf{Qualitative comparisons with regional prompt generated features and independently generated features from $2\times2$ splits.}}
    \vspace{-0.3cm}
\end{figure}

Here we propose to extract multiple features tailored to different aspects of an image, leveraging the instruction-following abilities of MLLMs. We observed that Regional Prompts (ReP), such as ``\texttt{<image> Summarize the [regional part] of the image in one word:}'', tend to focus on summarizing specific regions of the image. 

To show this, we begin by qualitatively analyzing~\cref{fig:directional-a}, which compares the cosine similarity between features generated using regional prompts ($f_A, f_B, f_C, f_D$) and features independently generated from A, B, C, D splits ($F_A, F_B, F_C, F_D$). The results reveal that $f$ closely aligns with $F$ for corresponding regions. Sequentially, in~\cref{fig:directional-b}, we compare $f_{A, B, C, D}$ and $F_{A, B, C, D}$ with the global image feature $F_{Global}$, showing that $f_{A, B, C, D}$ is consistently closer to the global image feature than $F_{A, B, C, D}$. Taking this analysis further, we randomly select 20 images from our benchmark and visualize a t-SNE map of independently generated patch split features and regional prompt-generated features. \cref{fig:directional-c} demonstrates that regional prompt-generated features cluster closely around the global image feature, while patch split features are often farther away, sometimes at significant distances. These observations demonstrate that regional prompt-generated features correspond to independently generated features of their respective regions, while retaining global information.

\subsection{Contrastive Finetuning}
\label{sec:finetune}

To enhance the discriminability of MLLM embeddings, previous works~\cite{E5-V, lamra, mm-embed, gme, discrim} adopt contrastive fine-tuning widely. Following this, we provide a simple but effective solution, by integrating Regional Prompts into the fine-tuning process. 

To employ regional prompts in fine-tuning, we restructure the dataset to include descriptions of each region. Using InternVL2.5-38B~\cite{internvl2.5}, we recaption the dataset by prompting it (see~\cref{supp:recaption_prompt}) to describe specific parts of the input image (\textit{left upper}, \textit{right upper}, \textit{left lower} and \textit{right lower}). This allows us to align prompt-generated features ($f_{lu}, f_{ru}, f_{ll}, f_{rl}$) with their corresponding text features ($t_{lu}, t_{ru}, t_{ll}, t_{rl}$). The text features are extracted using the prompt: ``\texttt{<text> Summarize the above sentence in one word:}''. To prevent redundancy from the same image appearing multiple times, we randomly select one regional prompt per image in training process. The training objective is as follows:

\begin{figure*}[t]
    \centering
    \includegraphics[width=1.0\linewidth]{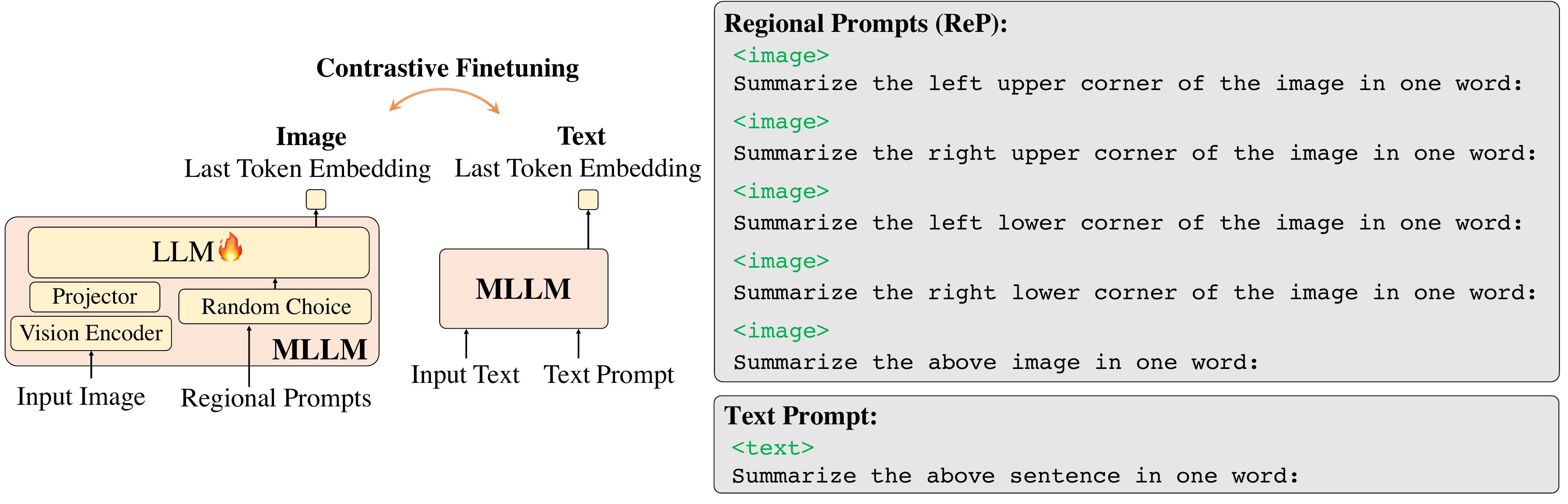}
    \caption{\textbf{Pipeline of our contrastive finetuning process.} \textit{Random Choice} block means that we randomly choose a regional prompt from the right, and use the resulting image embedding to align with the text description for the corresponding regional regions.}
    \vspace{-0.5cm}
    \label{fig:pipeline}
\end{figure*}

\begin{align}
    L = \frac{1}{2}\left(L_{CE}(f, t) + L_{CE}(t, f)\right),
\end{align}
where $L_{CE}$ is cross entropy loss and $L_{CE}(a, b)$ is:
\begin{align}
    L_{CE}(a, b) = -\log\frac{\exp \left(\cos(a^i, b^i)/\tau\right)}{\sum_j \exp(\cos(a^i, b^j)/\tau)}.
\end{align}
$\cos(\cdot, \cdot)$ denotes the cosine similarity function and $\tau$ is the temperature hyper-parameter. 

\subsection{Inference Setting}

Since retrieving a small target in a complex scene is agnostic to the query content, the generated image features must encompass as much information as possible to facilitate effective querying. Leveraging the text-guided feature extraction capability of MLLMs, we propose to use multiple features rather than a single feature to represent each image. Specifically, for each image, we generate five distinct features using five regional prompts, as listed in~\cref{fig:pipeline}. For a given query, we compute the cosine similarity between the query and each of the five features, selecting the closest one as the final feature for metric computation.

\section{Experiments}

\begin{table*}
\centering
\caption{\textbf{Image retrieval results on SORCE-1K.}
We report R@1, R@5, and R@10. The results indicate that ReP enhances the performance of MLLM-based feature extractors, while fine-tuning further improves retrieval recall. ft. means fine-tuned. Please refer to \cref{sec:finetune}.}
\label{tab:sorce}
\begin{adjustbox}{width=0.8\linewidth}
\begin{tabular}{ccccccccccc} 
\toprule
                         & \multirow{3}{*}{\begin{tabular}[c]{@{}c@{}}Multiple\\Features\end{tabular}} & \multicolumn{9}{c}{SORCE-1K (Ours)}                                                 \\ 
\cmidrule{3-11}
Method                   &                                                                          & \multicolumn{3}{c}{Zoom-$3\times$} & \multicolumn{3}{c}{Zoom-$2\times$} & \multicolumn{3}{c}{Full Res.}  \\ 
\cmidrule(lr){3-5}\cmidrule(lr){6-8}\cmidrule(lr){9-11}
\multicolumn{1}{l}{}     &                                                                          & R@1  & R@5  & R@10       & R@1  & R@5  & R@10         & R@1  & R@5  & R@10        \\ 
\midrule
CLIP ViT-B~\cite{clip}               & \ding{55}                                                                & 42.3 & 61.5 & 69.0       & 31.3 & 50.2 & 57.6         & 17.9 & 32.1 & 39.4        \\
CLIP ViT-L~\cite{clip}               & \ding{55}                                                                & 47.8 & 66.7 & 73.2       & 36.5 & 54.1 & 62.5         & 19.5 & 33.7 & 41.5        \\
EVA-02-CLIP 5B~\cite{EVA}    & \ding{55}    
       & 64.8 & 78.8 & 84.1      & 44.3 & 60.8 & 69.2  & 23.0 & 36.7 & 44.8 \\ 
LLaVA-Next-8B~\cite{llava-8b}               & \ding{55}                                                                & 41.0 & 59.1 & 69.1       & 28.2 & 44.6 & 52.4         & 13.4 & 23.0 & 30.4        \\
E5-V~\cite{E5-V}                     & \ding{55}                                                                & 57.6 & 74.0 & 82.2       & 42.4 & 62.2 & 69.0         & 21.9 & 36.7 & 44.6        \\ 
\midrule
E5-V + ReP    & \ding{51}                                                                & 62.0 & 80.6 & 86.3       & 54.0 & 73.2 & 80.2         & 27.7 & 45.4 & 53.7        \\
E5-V (ft.) + ReP & \ding{51}                                                                & \textbf{68.0} & \textbf{85.5} & \textbf{90.9}       & \textbf{56.3} & \textbf{77.1} & \textbf{83.0}          & \textbf{31.5} & \textbf{50.6} & \textbf{60.0}        \\
\bottomrule
\end{tabular}
\end{adjustbox}
\end{table*}


\begin{table*}
\centering
\caption{\textbf{Image retrieval results on Flickr30K~\cite{flickr} and COCO~\cite{coco}}. We also list text retrieval results for reference. The results indicate that MLLM-based feature extractors have comparable performance with their baseline model. ReP represents Regional Prompts.}
\label{tab:coco}
\begin{adjustbox}{width=0.9\linewidth}
\begin{tabular}{cccccccccccccc} 
\toprule
\multirow{3}{*}{Method}        & \multirow{3}{*}{\begin{tabular}[c]{@{}c@{}}Multiple\\Features\end{tabular}} & \multicolumn{6}{c}{Image Retrieval}                      & \multicolumn{6}{c}{Text Retrieval}                        \\ 
\cmidrule{3-14}
                               &                                                                          & \multicolumn{3}{c}{Flickr30K} & \multicolumn{3}{c}{COCO} & \multicolumn{3}{c}{Flickr30K} & \multicolumn{3}{c}{COCO}  \\ 
\cmidrule(lr){3-5} \cmidrule(lr){6-8} \cmidrule(lr){9-11} \cmidrule(lr){12-14}
                               &                                                                          & R@1  & R@5  & R@10            & R@1  & R@5  & R@10       & R@1  & R@5  & R@10            & R@1  & R@5  & R@10        \\ 
\midrule
CLIP ViT-B~\cite{clip}                    & \ding{55}                                                                & 62.1 & 85.6 & 91.8            & 33.1 & 58.4 & 69.1       & 81.9 & 96.2 & 98.8            & 52.5 & 76.7 & 84.6        \\
CLIP ViT-L~\cite{clip}                     & \ding{55}                                                                & 67.3 & 89.0 & 93.3            & 37.1 & 61.6 & 71.5       & 87.4 & 98.3 & 99.3            & 57.9 & 81.2 & 87.8        \\
LLaVA-Next-8B~\cite{llava-8b}                  & \ding{55}                                                                & 59.9 & 83.5 & 90.2            & 34.5 & 60.2 & 70.9       & 69.4 & 90.7 & 95.0            & 41.9 & 67.1 & 77.1        \\
EVA-02-CLIP 5B~\cite{EVA}                 & \ding{55}                                                                & 78.8 & 94.2 & 96.8            & 51.1 & 75.0   & 82.7       & \textbf{93.9} & \textbf{99.4} & \textbf{99.8}            & \textbf{68.8} & \textbf{87.8} & \textbf{92.8}        \\
E5-V~\cite{E5-V}                           & \ding{55}                                                                & 80.8 & 95.5 & 97.7            & \textbf{52.1} & \textbf{76.6} & 83.6       & 88.1 & 98.8 & 99.4            & 62.2 & 83.6 & 89.9        \\ 
\midrule
LLaVA-Next-8B + ReP & \ding{51}                                                                & 59.6 & 83.4 & 90.2            & 34.3 & 60.1 & 70.8       & 68.8 & 90.6 & 94.9            & 41.1 & 66.4 & 76.6        \\
E5-V + ReP          & \ding{51}                                                                & \textbf{81.0} & \textbf{95.7} & \textbf{97.8}            & 51.8 & 76.3 & \textbf{84.4}       & 87.9 & 98.7 & 99.4            & 59.9 & 82.0 & 88.6        \\
\bottomrule
\end{tabular}
\end{adjustbox}
\vspace{-0.5cm}
\end{table*}
\subsection{Experiment Settings}
\textbf{Datasets and benchmarks.} We use COCO-118K~\cite{coco, llava-next} for MLLM fine-tuning, which consists of 118K image-text pairs. We recaption the dataset with InternVL2.5-38B~\cite{internvl2.5}, with detailed prompting instructions provided in \cref{supp:recaption_prompt}. After recaptioning, each image in COCO-118K is paired with four regional descriptions (\textit{left upper corner, right upper corner, left lower corner, right lower corner}) and a summary caption. 

For evaluation, we mainly evaluate our method on commonly used retrieval benchmarks, Flickr30K~\cite{flickr} and COCO~\cite{coco} for validating the versatility of the proposed method in full image retrieval scenarios. In addition, we use the proposed benchmark to investigate the small object retrieval ability in complex environments, and set different difficulty levels, Zoom-$3\times$, Zoom-$2\times$ and Full Res., with increasing levels of challenge.

\label{sec:exp_detail}
\noindent \textbf{Implementation details.} Our method is trained on E5-V, which is pretrained on LLaVA-Next-8B~\cite{llava-8b}. From another perspective, our method can be seen as a two-stage training, which is first pretrained on text-only datasets, and then finetuned on text-image pairs, sharing the same intuition with the training strategy of LamRA~\cite{lamra}. We fine-tune the LLM part of the MLLM by QLoRA~\cite{qlora}, while keeping the image encoder and image feature projector frozen. 
We trained the model for a single epoch using 32 V100 GPUs, a batch size of 768, and a learning rate of $2\times 10^{-4}$, requiring approximately 3.5 hours.

\subsection{Comparison with Other Methods}
\label{sec:eval_detail}
The baselines to compare include CLIP with ViT-B and ViT-L-336px, and LLaVA-NeXT-8B, E5-V in MLLMs. For MLLMs, we use ``\texttt{<text> Summarize the above sentence in one word:}'' as the prompt for text feature extraction. For the proposed method, we extract five prompt-generated features for each image, while for other methods, each image is represented with one image feature.

We report Recall@K (R@K) with K=1, 5, 10 in~\cref{tab:coco} for image and text retrieval. For our benchmark in~\cref{tab:sorce}, we report image retrieval R@K with K=1, 5, 10. Compared to the baselines, our method keeps comparable performance on COCO and Flickr. And our method achieves competitive performance in our benchmark. Combining the performance on Flickr and COCO, our method showcases the potential of text-guided feature representation.

\subsection{Ablation Study}

\noindent \textbf{Does simple ensembling achieve the same performance?}
In the proposed method, we leverage multiple features, each focusing on different aspects of the image, to better capture its fine-grained details. A natural question arises: Does the performance improvement stem solely from ensembling multiple MLLM features? To investigate this, we use four additional synonyms for ``\texttt{Summarize}'' as global feature extraction prompts: ``\texttt{Conclude}'', ``\texttt{Synopsize}'', ``\texttt{Condense}'', and ``\texttt{Encapsulate}''. As shown in~\cref{tab:ensemble}, these five prompts yield negligible improvement over the single-feature setting. This demonstrates that our multiple regional prompts genuinely alter the focus of the MLLM-generated image features, rather than merely benefiting from ensembling.

\begin{table}
\caption{\textbf{Performance comparison with synonym prompts ensembling.}}
\label{tab:ensemble}
\centering
\begin{adjustbox}{width=0.6\linewidth}
\begin{tabular}{ccccc} 
\toprule
\multirow{2}{*}{Methods} & \multirow{2}{*}{\begin{tabular}[c]{@{}c@{}}Multiple\\Features\end{tabular}} & \multicolumn{3}{c}{SORCE-1K (R@5)}  \\ 
\cline{3-5}
                         &                                & Zoom-$3\times$ & Zoom-$2\times$ & Full Res.        \\ 
\midrule
E5-V                     & \ding{55}                      & 74.0    & 62.2    & 36.7      \\
+ Synonym Prompts   & \ding{51}                      & 74.5    & 61.8    & 36.5      \\
+ ReP   & \ding{51}                      & \textbf{85.5}    & \textbf{77.1}    & \textbf{50.6}      \\
\bottomrule
\end{tabular}
\end{adjustbox}
\end{table}

\noindent \textbf{Why not obtain multiple features by cropping?}
Cropping is an intuitive approach to generating multiple features from an image. However, it has significant limitations: it lacks global context when extracting features from each cropped region and risks splitting the target object entirely. Here we crop the image into $2\times 2$ non-overlapping splits and extract features from each region. We also append the global feature when evaluating, resulting in a set of five features per image and ensuring a fair comparison. In~\cref{tab:crop-ours}, cropping improves performance over the base model E5-V and surpasses regional prompts. However, this is largely due to our benchmark only involving minimal required context for identifying the target small object. In contrast, \cref{tab:crop-coco} shows that cropping significantly degrades performance, indicating that it is not a universally applicable solution.

\begin{table}[t]
    \centering
    \begin{minipage}{0.49\textwidth}
        \centering
        \caption{\textbf{Performance comparison between $2\times 2$ split features and regional prompt features on SORCE-1K.} }
        \label{tab:crop-ours}
        \adjustbox{max width=\linewidth}{
            \begin{tabular}{ccccc} 
            \toprule
            \multirow{2}{*}{Methods} & \multirow{2}{*}{\begin{tabular}[c]{@{}c@{}}Multiple\\Features\end{tabular}} & \multicolumn{3}{c}{SORCE-1K (R@5)}   \\ 
            \cline{3-5}
                                     &                                                                         & Zoom 3$\times$ & Zoom 2$\times$ & Full Res.  \\ 
            \midrule
            E5-V                     & \ding{55}                                                               & 74.0    & 62.2    & 36.7       \\
            + $2\times 2$ Split              & \ding{51}                                                               & 79.4    & 70.8    & 47.9       \\
            + ReP        & \ding{51}                                                               & 80.6    & 73.2    & 45.4       \\
            (ft.) + ReP     & \ding{51}                                                               & \textbf{85.5}    & \textbf{77.1 }   & \textbf{50.6}       \\
            \bottomrule
        \end{tabular}
        }

    \end{minipage}
    \hfill
    \begin{minipage}{0.4\textwidth}
        \centering
        \caption{\textbf{Performance comparison between $2\times 2$ split features and regional prompt features on Flickr and COCO.} }
\label{tab:crop-coco}
\adjustbox{max width=\linewidth}{
        \begin{tabular}{cccc} 
            \toprule
            \multirow{2}{*}{Methods} & \multirow{2}{*}{\begin{tabular}[c]{@{}c@{}}Multiple\\Features\end{tabular}} & Flickr30K & COCO   \\ 
            \cmidrule(lr){3-3} \cmidrule(lr){4-4}
                                     &                                                                         & R@5       & R@5    \\ 
            \midrule
            E5-V                     & \ding{55}                                                               & 95.5      & \textbf{76.6}   \\
            + $2\times2$ Split              & \ding{51}                                                               & 84.8     & 59.4  \\
            + ReP        & \ding{51}                                                               & \textbf{95.7}      & 76.3   \\
            \bottomrule
        \end{tabular}}
    \end{minipage}

    \centering
    \begin{minipage}{0.45\textwidth}
        \centering
        \caption{\textbf{Effects of Different Prompt Numbers.} In sequence, $P_a, P_b, P_c, P_d, P_e$ represent four regional prompts followed by one summary prompt. }
        \label{tab:prompt-num}
        \adjustbox{max width=\linewidth}{
            \begin{tabular}{cccc} 
            \toprule
            \multirow{2}{*}{Methods} & \multicolumn{3}{c}{SORCE-1K (R@5)}  \\ 
            \cline{2-4}
                                     & Zoom-3$\times$ & Zoom-2$\times$ & Full Res.       \\ 
            \hline
            E5-V                     & 74.0    & 62.2    & 36.7            \\
            $+P_a$                      & 64.4   & 54.0   & 32.0           \\
            $+P_{a, b}$                    & 65.9   & 56.7    & 36.5           \\
            $+P_{a, b, c}$                & 73.6   & 67.7   & 39.3            \\
            $+P_{a, b, c, d}$                & 75.2   & 69.8   & 42.2           \\
            $+P_{a, b, c, d, e}$              & \textbf{80.6}    & \textbf{73.2}    & \textbf{45.4}            \\
            \bottomrule
            \end{tabular}
        }
    \end{minipage}
    \hfill
    \begin{minipage}{0.5\textwidth}
        \centering
        \caption{\textbf{Comparison between directly contrastive finetuning and random choice finetuning.}
        M-All means merge the regional descriptions of each recaptioned image into one caption, and perform contrastive fine-tuning.}
        \label{tab:merge-all}
        \adjustbox{max width=\linewidth}{
            \begin{tabular}{ccccc} 
            \toprule
            \multirow{2}{*}{Methods} & \multirow{2}{*}{\begin{tabular}[c]{@{}c@{}}Multiple\\Features\end{tabular}} & \multicolumn{3}{c}{SORCE-1K (R@5)}  \\ 
            \cmidrule{3-5}
                                     &                                                                         & Zoom 3-$\times$ & Zoom 2-$\times$ & Full Res.       \\ 
            \midrule
            E5-V                     & \ding{55}                                                               & 74.0    & 62.2    & 36.7            \\
            + ReP        & \ding{51}                                                               & 80.6    & 73.2    & 45.4            \\
            (M-All ft.) + ReP           & \ding{51}                                                               & 80.3    & 69.9    & 43.1            \\
            (ft.) + ReP     & \ding{51}                                                               & \textbf{85.5}    & \textbf{77.1}    & \textbf{50.6}            \\
            \bottomrule
            \end{tabular}        
        }
    \end{minipage}

\end{table}

\noindent \textbf{Does the number of the prompts affect performance?}
We denote the left upper corner, right upper corner, left lower corner, right lower corner and global summary as $a, b, c, d, e$ accordingly. Then we ablate with 1 to 5 prompts, from $a$ to $a, b, c, d, e$, denoting as $P_a, P_{a, b}, ..., P_{a, b, c, d, e}$. From~\cref{tab:prompt-num}, we can see that with more prompt-generated features included, the performance keeps improving.


\noindent \textbf{Does the regional prompt finetuning truly improve performance?} 
We transform our InternVL2.5-38B recaptioned COCO-118K dataset into a format where each image is associated with a single unified description, and apply the contrastive tuning between the global image feature and the full description. As shown in~\cref{tab:merge-all}, the instruction-following capability exhibits a slight decline after fine-tuning, suggesting that our random choice fine-tuning strategy provides a beneficial effect.

We present more ablation study like the effect of using different prompts in~\cref{supp:other_prompts}.
\vspace{-0.2cm}

\section{Conclusion}
\vspace{-0.2cm}
In this work, we focus on an overlooked demand in T2IR, which is retrieving a specific small object from images with complex scenes.
We start by formulating the corresponding task setting, Small Object Retrieval in Complex Environments, and constructing the SORCE-1K benchmark. The benchmark focuses on retrieving small objects from large images, based on language descriptions, without the interference of contexts.
Noticing that a single feature may not represent all small objects in an image, we leverage MLLMs to extract text-custimized features focusing on various aspects in an image, which provides a simple and intuitive solution for this setting.
We hope the SORCE-1K benchmark, together with the proposed solution, can serve as a foundation for future research on such a setting.
We discuss the limitations and broader impact in~\cref{supp:limitation,supp:impact}.


    \bibliographystyle{ieeenat_fullname}

\bibliography{main}


\appendix

\clearpage
\appendix

\section*{Appendix}

\section{Environment Complexity of SORCE-1K}
\label{supp:environment_complexity}

\begin{figure}[h]
    \centering
    \includegraphics[width=0.45\linewidth]{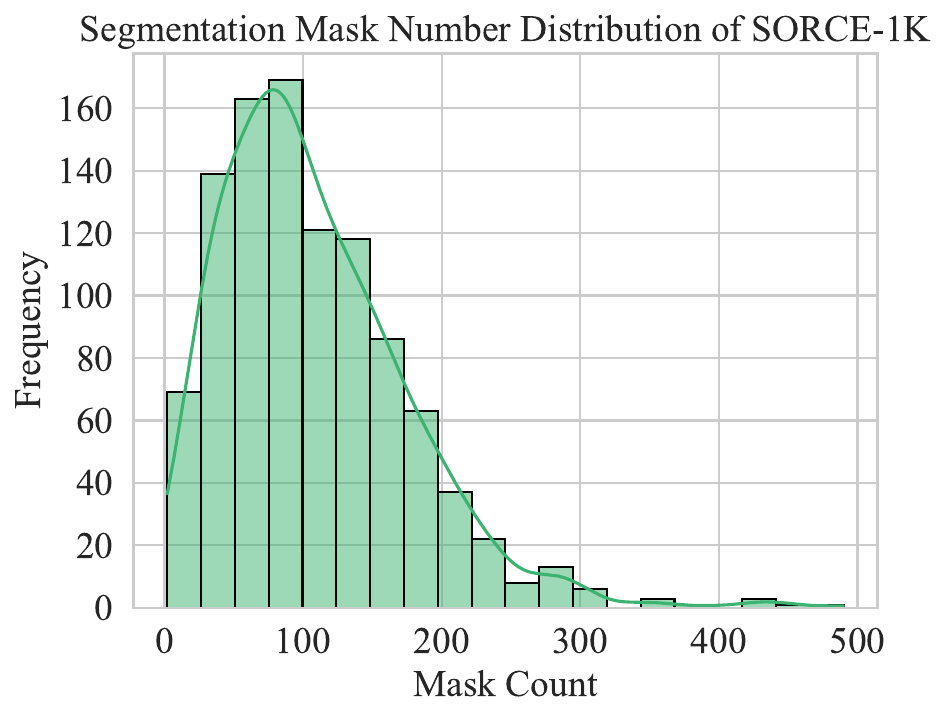}
    \caption{\textbf{Mask distribution of SORCE-1K.} Environment complexity usually grows with the number of segmentation masks.}
    \label{fig:masks}
\end{figure}

Since our SORCE-1K benchmark is collected from SA-1B~\cite{sam}, every image has the corresponding mask label. The complexity of the image usually increases with the number of segmentation masks. Therefore, we provide the segmentation mask number distribution of SORCE-1K in~\cref{fig:masks}.

\section{The Caption Length Distribution of Flickr30K and COCO}
\label{supp:caption_length}

\begin{figure}
    \centering
    \begin{subfigure}{0.48\linewidth}
        \includegraphics[width=0.9\linewidth]{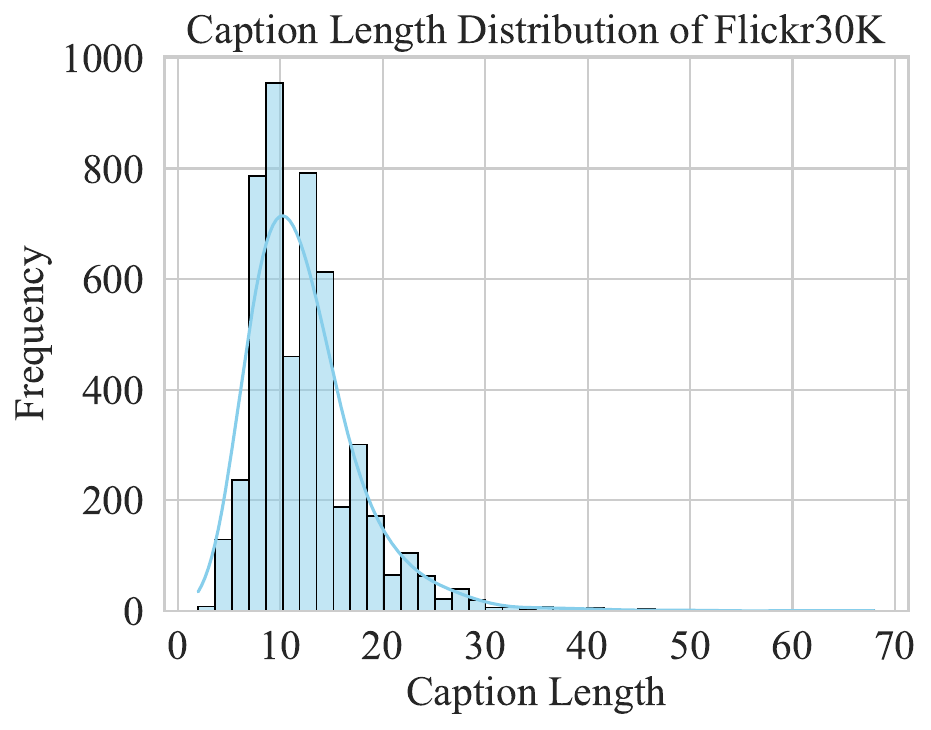}
        \caption{Flickr30K}
    \end{subfigure}
    \begin{subfigure}{0.48\linewidth}
        \includegraphics[width=0.9\linewidth]{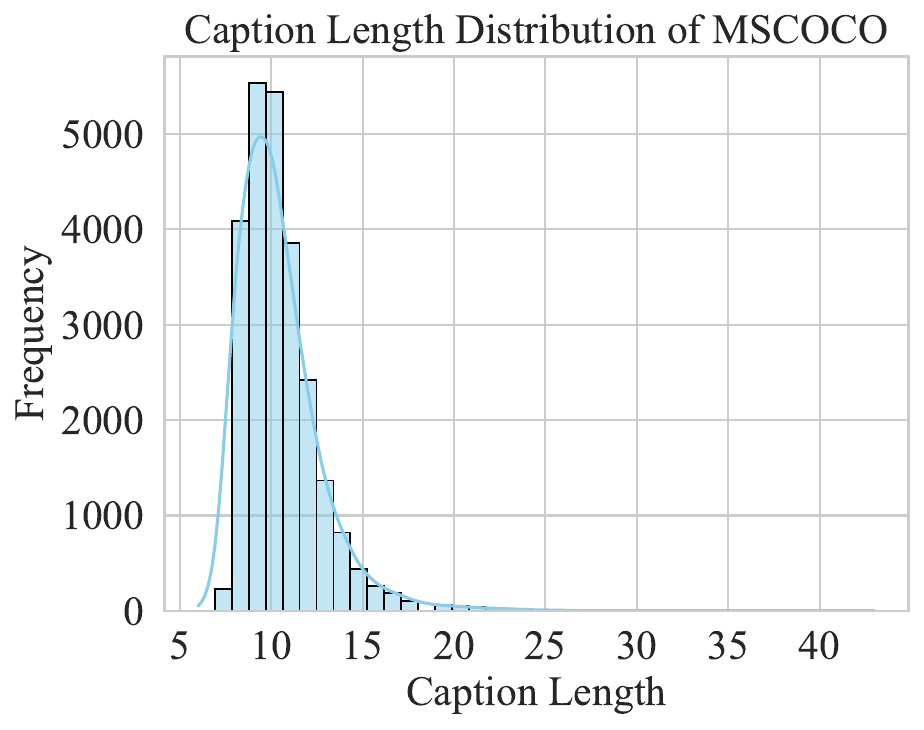}
    \caption{COCO}
    \end{subfigure}
    \caption{\textbf{Caption length distribution charts of Flickr30K and COCO}.}
    \label{fig:cap-len}
\end{figure}

We present the caption length distribution charts in~\cref{fig:cap-len}, alongside the SORCE-1K caption length distribution in~\cref{fig:sorce-stats}. As shown, the caption lengths in SORCE-1K are comparable to or slightly longer than those in COCO and Flickr. However, the key distinction lies in the focus of the captions: SORCE-1K emphasizes detailed descriptions of the target object while minimizing the inclusion of global context. Consequently, caption length is not the primary metric we aim to highlight.

\section{Construction Details of SORCE-1K}
\label{supp:construction_details}

Here we provide more details regarding the construction process (see~\cref{sec:benchmark}) of the proposed SORCE-1K benchmark.

\noindent \textbf{Region-description pair annotation.}
This step is performed manually by researchers ourselves. We use the online annotation tool\footnote{\url{https://trexlabel.com/}} as the interface to draw the bounding boxes, and design the descriptions manually. In the annotation process, we try to select small objects taking up less than 10\% of the image. However, there are no explicit numerical constraints in this step, as we can filter them by the area ratio to the whole image later.

\noindent \textbf{Automatic check.}
We use Python scripts to automatically remove the samples where the bounding box of the target object takes up more than 20\% of the image; we also automatically pick out the samples whose corresponding query descriptions are shorter than 8 words, and extend their lengths to at least 8 words.

\section{More Visualization Examples in SORCE-1K}
\label{supp:more_qualitative}

\begin{figure*}
    \centering
    \begin{subfigure}{0.9\linewidth}
        \includegraphics[width=1.0\linewidth]{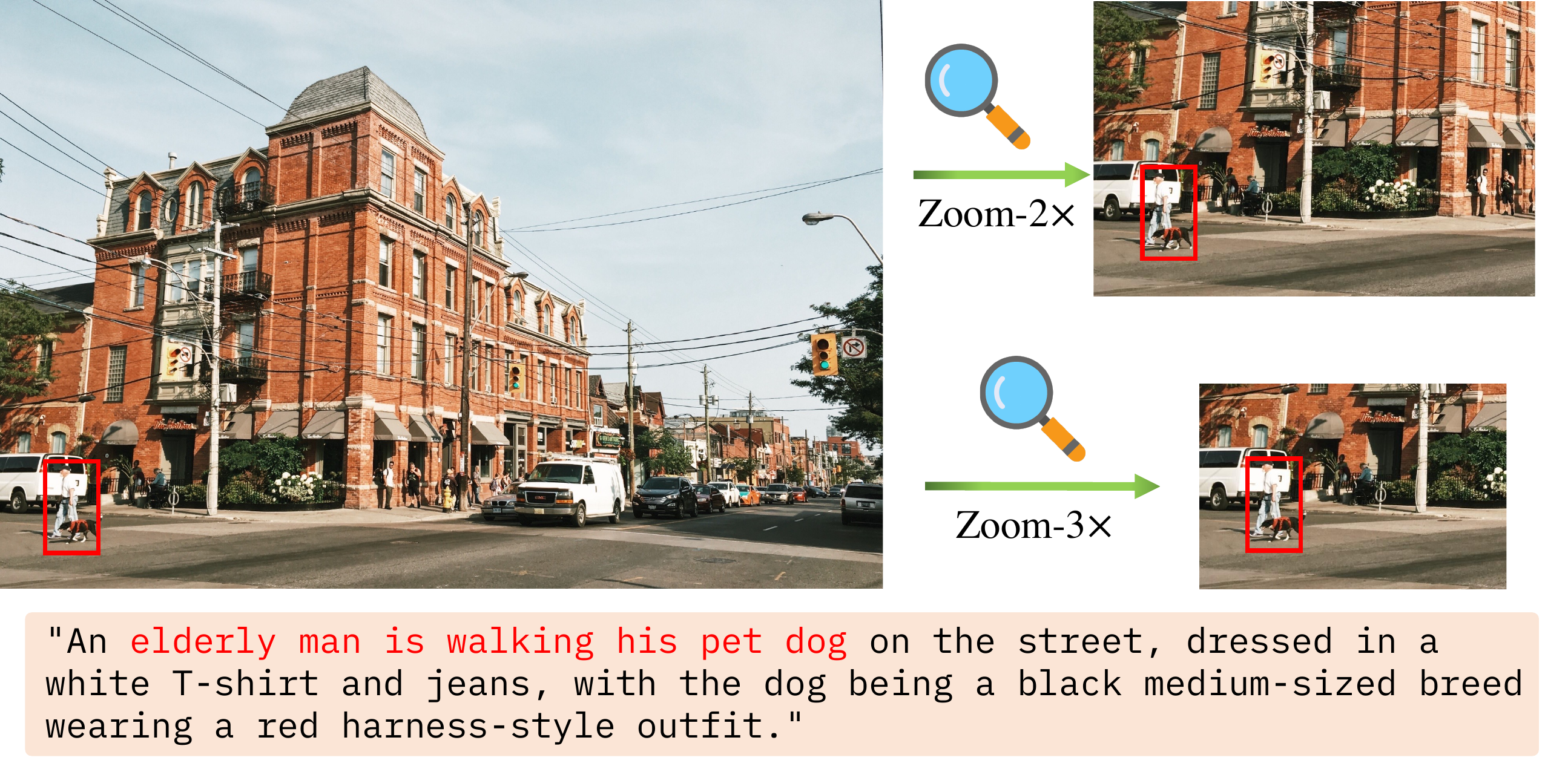}
    \end{subfigure}
    \vspace{0.5cm}
    
    \begin{subfigure}{0.9\linewidth}
        \includegraphics[width=1.0\linewidth]{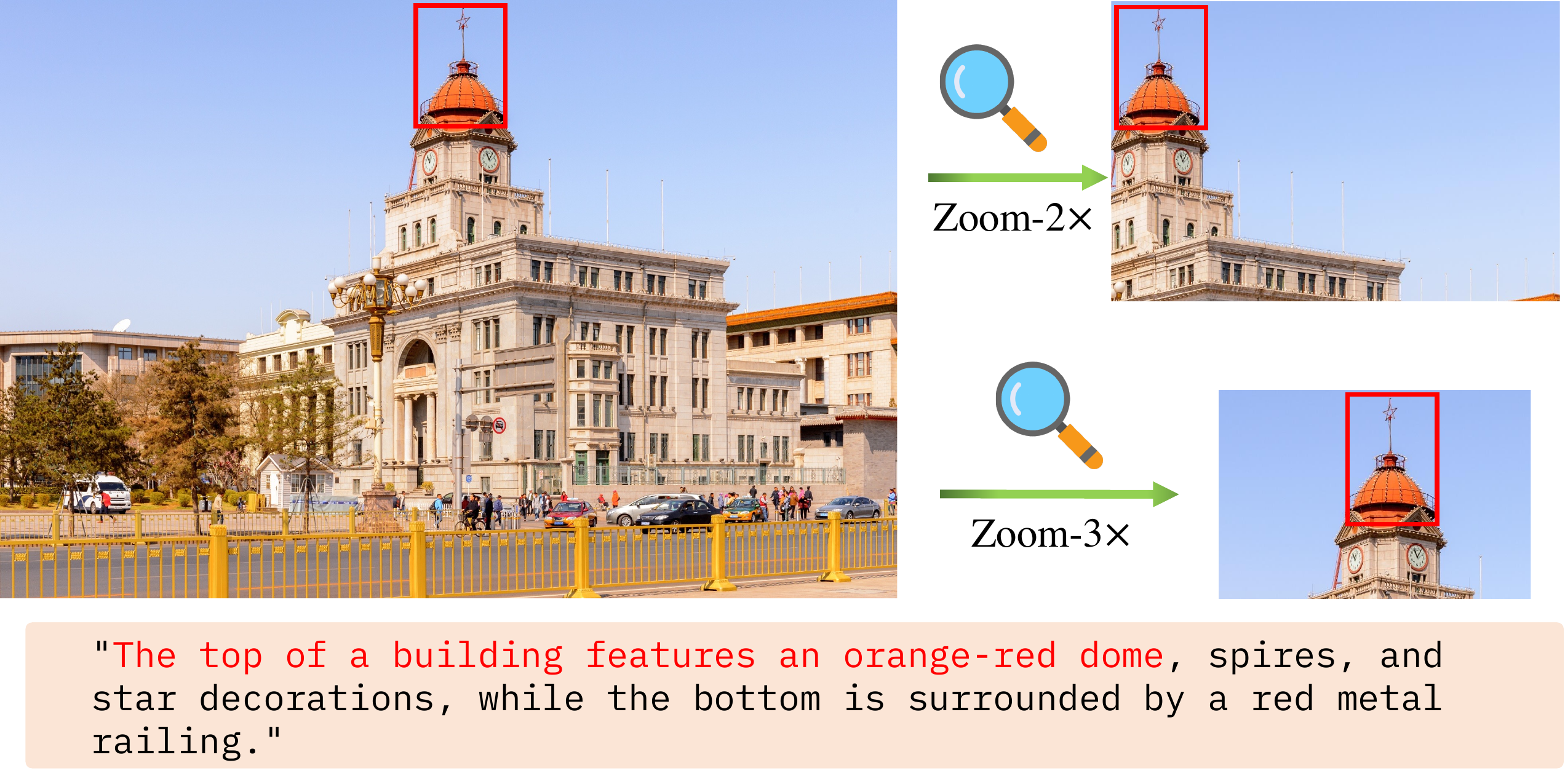}
    \end{subfigure}
    \vspace{0.5cm}
    
    \begin{subfigure}{0.9\linewidth}
        \includegraphics[width=1.0\linewidth]{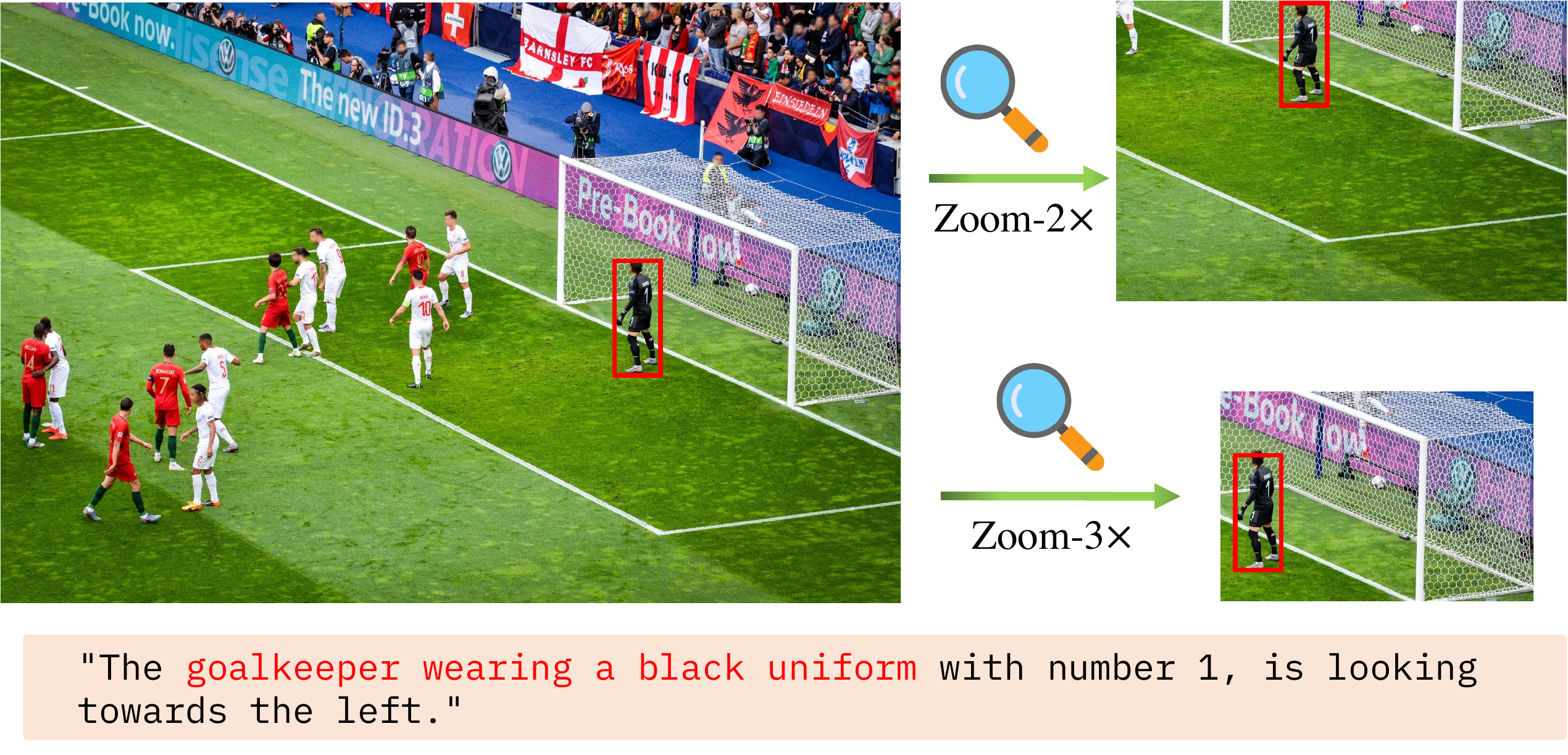}
    \end{subfigure}
    
    \caption{\textbf{More visualization examples from SORCE-1K benchmark.}}
    \label{fig:more-vis}
\end{figure*}

In~\cref{fig:more-vis}, we present more samples from our benchmark for reference.

\section{Re-caption Prompts for COCO-118K}
\label{supp:recaption_prompt}

The prompt we use for guiding InternVL2.5-38B~\cite{internvl2.5} is as follows: 

\texttt{Describe the image in a structured manner, following this order: left upper, right upper, left lower, right lower, and summary. For each area, include detailed descriptions of key elements, such as objects, colors, textures, sizes, actions, or any other relevant features. Ensure each description is under 100 words, but provide enough detail to give a vivid picture. The output should be a dictionary with the following keys: `left upper', `right upper', `left lower', `right lower' and `summary.'}

The resulting average length of generated captions is listed in~\cref{tab:avg-len}

\begin{table}
\centering
\caption{\textbf{Average sentence length of generated captions of COCO-118K by InternVL2.5-38B.} Length is measured in words.}
\label{tab:avg-len}
\begin{tabular}{cc} 
\toprule
            & Average Length (Words)  \\ 
\midrule
Left Upper  & 37.7                    \\
Right Upper & 36.8                    \\
Left Lower  & 36.3                    \\
Right Lower & 48.2                    \\
Summary     & 30.8                    \\ 
\midrule
All         & 35.5                    \\
\bottomrule
\end{tabular}
\end{table}

\section{Exploration of Other Prompts}
\label{supp:other_prompts}

\begin{table*}
\centering
\caption{\textbf{Image retrieval results on Flickr30K~\cite{flickr} and COCO~\cite{coco}}. We also list text retrieval results for reference. The results indicate that MLLM-based feature extractors have comparable performance to their baseline model. ReP represents Regional Prompts.}
\label{tab:coco-sup}
\begin{adjustbox}{width=1.0\linewidth}
\begin{tabular}{cccccccccccccc} 
\toprule
\multirow{3}{*}{Method}        & \multirow{3}{*}{\begin{tabular}[c]{@{}c@{}}Multiple\\Features\end{tabular}} & \multicolumn{6}{c}{Image Retrieval}                      & \multicolumn{6}{c}{Text Retrieval}                        \\ 
\cmidrule{3-14}
                               &                                                                          & \multicolumn{3}{c}{Flickr30k} & \multicolumn{3}{c}{COCO} & \multicolumn{3}{c}{Flickr30k} & \multicolumn{3}{c}{COCO}  \\ 
\cmidrule(lr){3-5} \cmidrule(lr){6-8} \cmidrule(lr){9-11} \cmidrule(lr){12-14}
                               &                                                                          & R@1  & R@5  & R@10            & R@1  & R@5  & R@10       & R@1  & R@5  & R@10            & R@1  & R@5  & R@10        \\ 
\midrule
E5-V~\cite{E5-V}                           & \ding{55}                                                                & 80.8 & 95.5 & 97.7            & 52.1 & 76.6 & 83.6       & 88.1 & 98.8 & 99.4            & 62.2 & 83.6 & 89.9        \\ 
\midrule

E5-V + Semantic Prompts          & \ding{51}                                                                & 80.7 & 95.5 & 97.6            & \textbf{52.8} & \textbf{77.1} & \textbf{85.2 }      & 86.8 & 98.1 & 99.0            & 59.6 & 82.4 & 88.8        \\
E5-V + ReP          & \ding{51}                                                                & 81.0 & 95.7 & 97.8            & 51.8 & 76.3 & 84.4       & 87.9 & 98.7 & 99.4            & 59.9 & 82.0 & 88.6        \\
E5-V (ft.) + Semantic Prompts          & \ding{51}                                                       
&\textbf{81.1}	&\textbf{95.8}	&97.9	&51.8	&76.4	&84.5	&91.9	&\textbf{99.1}	&99.7	&65.1	&\textbf{86.7}	&\textbf{92.5} \\
E5-V (ft.) + ReP          & \ding{51}                                                                &80.7	&95.7	&\textbf{98.0}	&50.6	&75.6	&83.7	&\textbf{93.8}	&\textbf{99.1}	&\textbf{99.8}	&\textbf{65.5}	&86.4	&92.0 \\
\bottomrule
\end{tabular}
\end{adjustbox}
\end{table*}

\begin{table*}
\centering
\caption{\textbf{Image retrieval results on SORCE-1K.} We report R@1, R@5, and R@10. The results indicate that ReP enhances the performance of MLLM-based feature extractors, while fine-tuning further improves retrieval recall. (ft.) means fine-tuned, please refer to Section \ref{sec:finetune}. ReP represents Regional Prompts.}
\label{tab:sorce-sup}
\begin{adjustbox}{width=0.8\linewidth}
\begin{tabular}{ccccccccccc} 
\toprule
                         & \multirow{3}{*}{\begin{tabular}[c]{@{}c@{}}Multiple\\Features\end{tabular}} & \multicolumn{9}{c}{SORCE-1K (Ours)}                                                 \\ 
\cmidrule{3-11}
Method                   &                                                                          & \multicolumn{3}{c}{Zoom-$3\times$} & \multicolumn{3}{c}{Zoom-$2\times$} & \multicolumn{3}{c}{Full Res.}  \\ 
\cmidrule(lr){3-5}\cmidrule(lr){6-8}\cmidrule(lr){9-11}
\multicolumn{1}{l}{}     &                                                                          & R@1  & R@5  & R@10       & R@1  & R@5  & R@10         & R@1  & R@5  & R@10        \\ 
\midrule
E5-V~\cite{E5-V}                     & \ding{55}                                                                & 57.6 & 74.0 & 82.2       & 42.4 & 62.2 & 69.0         & 21.9 & 36.7 & 44.6        \\ 
\midrule

E5-V + Semantic Prompts    & \ding{51}                                                                & 57.9 & 74.2 & 81.8       & 42.4 & 61.1 & 68.9         & 21.0 & 36.1 & 45.1        \\
E5-V + ReP    & \ding{51}                                                                & 62.0 & 80.6 & 86.3       & 54.0 & 73.2 & 80.2         & 27.7 & 45.4 & 53.7        \\
E5-V (ft.) + Semantic Prompts    & \ding{51}                                                                
&65.8	&82.8	&88.1	&53.5	&72.4	&80.2	&28.0	&45.7	&54.6 \\
E5-V (ft.) + ReP & \ding{51}                                                                & \textbf{68.0} & \textbf{85.5} & \textbf{90.9}       & \textbf{56.3} & \textbf{77.1} & \textbf{83.0}          & \textbf{31.5} & \textbf{50.6} & \textbf{60.0}        \\
\bottomrule
\end{tabular}
\end{adjustbox}
\end{table*}

In addition to Regional Prompts (ReP), we also explored extracting multiple features using semantic prompts, such as ``\texttt{Summarize the [main component/ background/ detail] in the above image in one word:}''. We append the global feature to these, resulting in a set of four features per image. For training dataset construction, we prompt the InternVL2.5-38B to generate descriptions of each image from different perspectives. We then employ contrastive fine-tuning with random choice, using the same learning hyperparameters as described in~\cref{sec:finetune}.

As shown in~\cref{tab:coco-sup}, the performance of semantic prompts remains comparable after fine-tuning. However, in~\cref{tab:sorce-sup}, while semantic prompts also improve performance after fine-tuning, regional prompts consistently outperform them. We attribute this to the inherent ambiguity of semantic prompts, as it is often unclear which part of the image constitutes the "main component" or "detail." This conclusion is evident from the E5-V before finetuning, indicating that regional prompt instructions are easier for MLLMs to follow. Nevertheless, these results highlight the potential of MLLM-based text-guided feature representations for enhancing retrieval tasks.

\section{Discussions and Limitations.}
\label{supp:limitation}

Although extracting multiple features by regional prompting can keep the performance of the common image retrieval, and exhibits superior performance on small objects in complex scene image retrieval. The storage requirement also multiplies. Therefore, a feature extractor with the capability to preserve every piece of visual information into one feature is still an ideal option. However, this is very challenging. Therefore, as a compromise solution, MLLM as a feature extractor is worth discussing. For instance, we could see the potential through our paper, which is that text-guided feature extraction is possible. Further, we could possibly utilize the autoregressive ability of MLLM to adaptively extract important features for the image, saving the need for extracting a fixed number of features for each image.

\section{Broader Impact}
\label{supp:impact}

In this work, we introduce the task of Small Object Retrieval in Complex Environments, namely, SORCE, the first T2IR benchmark focused on unsalient small objects. And we propose a benchmark, SORCE-1K, to evaluate the performance of the models. Furthermore, we provide an initial solution by Regional Prompts (ReP) to extract corresponding image features which can focus on different aspects of the image while keeping the global information. The SORCE task and our benchmark are beneficial for enhancing the understanding and feature extraction performance of multimodal models. Also, as a research-oriented work and the initial attempt on such a task, we trained our model on a very limited set of dataset (COCO-118K~\cite{coco}), which is only for validating our idea and may not be ready for real-world applications. It may be mitigated with fine-tuning more real-world-related datasets. 

\section{License of datasets and pre-trained models}
\label{supp:license}

All the datasets we used in the paper are commonly used datasets for academic purposes. All the licenses of the used benchmark, codes, and pretrained models are listed in~\cref{tab:license}.

\begin{table}
\centering
\caption{Licenses and URLs for every dataset, benchmark, code, and pretrained models used in this paper.}
\label{tab:license}
\adjustbox{max width=\linewidth}{
\begin{tabular}{cccc} 
\toprule
\multicolumn{2}{c}{Assets}                                                                          & License                & URL                                                                                                    \\ 
\midrule
\multirow{3}{*}{\begin{tabular}[c]{@{}c@{}}Dataset and\\Benchmarks\end{tabular}}      & MSCOCO      & CC BY 4.0              & \textcolor{blue}{\textcolor[rgb]{0.09,0.361,0.922}{https://cocodataset.org/}}                          \\
                                                                                      & Flickr30K   & Research purposes only & \textcolor{blue}{\textcolor[rgb]{0.09,0.361,0.922}{https://shannon.cs.illinois.edu/DenotationGraph/}}  \\
                                                                                      & SA-1B       & Research purposes only & \textcolor{blue}{\textcolor[rgb]{0.09,0.361,0.922}{https://ai.meta.com/datasets/segment-anything/}}    \\ 
\midrule
\multirow{5}{*}{\begin{tabular}[c]{@{}c@{}}Codes and\\Pretrained Models\end{tabular}} & CLIP        & MIT-license            & \textcolor{blue}{\textcolor[rgb]{0.09,0.361,0.922}{https://github.com/openai/CLIP}}                    \\
                                                                                      & EVA-02-CLIP & MIT-license            & \textcolor{blue}{\textcolor[rgb]{0.09,0.361,0.922}{https://github.com/baaivision/EVA}}                 \\
                                                                                      & LLaVA-Next  & Apache-2.0 license     & \textcolor{blue}{\textcolor[rgb]{0.09,0.361,0.922}{https://github.com/LLaVA-VL/LLaVA-NeXT}}            \\
                                                                                      & E5-V        & Research purposes only & \textcolor{blue}{\textcolor[rgb]{0.09,0.361,0.922}{https://github.com/kongds/E5-V}}                    \\
                                                                                      & InternVL2.5 & MIT-license            & \textcolor{blue}{\textcolor[rgb]{0.09,0.361,0.922}{https://github.com/OpenGVLab/InternVL/tree/main}}   \\
\bottomrule
\end{tabular}
}
\end{table}



\end{document}